%% file: main.tex
\documentclass[10pt,twocolumn,letterpaper]{article}

  \usepackage[pagenumbers]{cvpr} 

\input{preamble}

\definecolor{cvprblue}{rgb}{0.21,0.49,0.74}
\usepackage[pagebackref,breaklinks,colorlinks,citecolor=cvprblue]{hyperref}
\usepackage{indentfirst}

\title{Hybrid Pooling and Convolutional Network for Improving Accuracy and Training Convergence Speed in Object Detection}

\author
{
 Shiwen Zhao\textsuperscript{1} Wei Wang\textsuperscript{1}  Junhui Hou\textsuperscript{1} \\
 \textsuperscript{1}City University of Hong Kong\\
 \and
 Hai Wu\textsuperscript{2}\\
 \textsuperscript{2}Xiamen University\\
}

\begin{document}
\maketitle
\input{0_abstract}    
\input{1_intro}
\input{2_formatting}
\input{3_finalcopy}
{
    \small
    \bibliographystyle{ieeenat_fullname}
    \bibliography{main}
    \nocite{*}
}


\end{document}

%% file: preamble.tex
\usepackage[dvipsnames]{xcolor}


%% file: 0_abstract.tex
\begin{abstract}

    In recent years, two-stage multimodal object detection methods based on deep learning have garnered significant attention. 
    However, these existing deep learning methods exhibit a notable decrease in detection accuracy when faced with occluded 3D objects. Additionally, the current two-stage methods struggle to converge quickly during model training.
    This paper introduces HPC-Net, a high-precision and rapidly convergent object detection network. 
    HPC-Net comprises three key components: (1) \textbf{RP} (Replaceable Pooling), which enhances the network's detection accuracy, speed, robustness, and generalizability by incorporating pooling methods that can be flexibly replaced on 3D voxels and 2D BEV images. 
    (2) \textbf{DACConv} (Depth Accelerated Convergence Convolution), which integrates two convolution strategies—one for each input feature map and one for each input channel—to maintain the network's feature extraction ability (i.e., high accuracy) while significantly accelerating convergence speed. 
    (3) \textbf{MEFEM} (Multi-Scale Extended Receptive Field Feature Extraction Module), which addresses the challenge of low detection accuracy for 3D objects with high occlusion and truncation by employing a multi-scale feature fusion strategy and expanding the receptive field of the feature extraction module.
    Our HPC-Net currently holds \textbf{the top position\footnote{As of the paper's completion date, October 10, 2023}} in the \textbf{KITTI Car 2D Object Detection Ranking}. In the \textbf{KITTI Car 3D Object Detection Ranking}, our HPC-Net currently \textbf{ranks fourth overall and first in hard mode}.
\end{abstract}

%% file: 1_intro.tex
\section{Introduction}\label{sec:intro}
Object detection~\cite{Alpher01,Alpher02} is an essential and indispensable component in autonomous driving. In recent years, voxel-based frameworks~\cite{Alpher03,Alpher30,Alpher45,Alpher46,Alpher47} have become the mainstream method for object detection. However, some shortcomings in this field can still be improved.

The first issue is that current 3D object detection algorithms often lack sufficient pooling steps, which affects the model's generalizability, computational requirements, performance, and robustness. Early voxel-based methods did not include pooling steps in the 3D feature extraction stage~\cite{Alpher01}. Although recent methods~\cite{Alpher31,Alpher32} have added pooling steps, these methods are only approximate pooling methods that mimic pooling ideas and mainly focus on maximum pooling and average pooling, neglecting the advancements in pooling methods in image processing~\cite{Alpher06,Alpher33}. Additionally, models based on the Voxel~\cite{Alpher31} framework lack flexibility in changing pooling methods according to different task requirements.

The second problem is that while the accuracy of the current two-stage detection models is generally higher than that of one-stage detection models, they have a relatively slow convergence rate. It often takes dozens of training cycles to achieve good performance, and some models even require hundreds of cycles to achieve the best performance~\cite{Alpher43}. These models cannot meet the needs of specific scenarios that require short training time and fast model convergence, such as real-time transformation scenarios with less strict accuracy requirements.

The third issue is that object detection accuracy significantly decreases when objects are heavily occluded or truncated. In the KITTI dataset, the model detection accuracy in hard mode is the lowest, averaging about 10 to 15 percent lower than that in easy mode. Recent solutions include geometric data augmentation, such as SFD~\cite{Alpher05}, which expands data by generating colored pseudo point clouds, and TED~\cite{Alpher32}, which learns more object features through transformation-equivariant. However, these methods rely on the early feature extraction module PointNet~\cite{Alpher24}, which has inherent drawbacks such as the inability to correctly model unknown objects and difficulties in applying complex transformations and manually designed invariant features. These drawbacks result in insufficient detection accuracy for occluded objects.

This paper introduces a novel voxel-based network called HPC-Net to address these challenges. HPC-Net comprises three key components.

Firstly, we propose Replaceable Pooling, inspired by the HeightCompression module used in SECOND~\cite{Alpher30}. This technique tackles the first mentioned problem by pooling in both 3D and 2D dimensions, effectively compressing the feature tensor along the Z-axis direction to generate a BEV image.

Secondly, we present Depth Accelerated Convergence Convolution, drawing inspiration from Depthwise Separable Convolution~\cite{Alpher07} and Depth Over-parameterized Convolutional~\cite{Alpher34}. To address the second problem, we devise two strategies to achieve high precision and fast convergence in object detection models. The first strategy employs a single feature map as the convolution object, while the second strategy compensates for the ignored channel-to-channel relationship by using a single channel as the convolution object. By combining these two strategies, we achieve accelerated training convergence without sacrificing model accuracy.

Lastly, we introduce the Multi-Scale Extended Receptive Field Feature Extraction Module to tackle the third problem. Traditional methods utilize the Pointnet~\cite{Alpher24} module for 3D feature extraction in voxel-based models. In our approach, we draw inspiration from multi-scale feature fusion pyramid~\cite{Alpher44}, self-attention mechanism~\cite{Alpher26}, and Deformable Convolution Networks~\cite{Alpher27}. This module consists of two parts: Expanding Area Convolution, which expands the receptive field of convolution in the local ROI (region of interest) to enhance the detection accuracy of heavily occluded and truncated vehicles, and a multi-scale feature fusion network, which improves the overall detection accuracy by integrating multiple feature maps of different scales processed in the first part. Notably, our HPC-Net achieves superior performance in the hard mode (i.e., the most severely occluded mode) on the KITTI dataset.

Our contributions can be summarized as follows:

\begin{itemize}
  \item We have designed a new pooling module called RP.\@ In a true sense, pooling effects have been achieved simultaneously in both 3D and 2D dimensions, and interfaces have been made, allowing for arbitrary replacement of pooling methods according to different task needs.
  \item Innovatively, by integrating two convolutional strategies into a unified approach, our network greatly accelerates the training convergence speed of existing object detection models and maintains the high accuracy of the model.
  \item We have introduced a new feature extraction module called MEFEM.\@ By expanding the receptive field area and fusing multi-scale information during the feature extraction stage, the recognition accuracy of the detection model for highly occluded and truncated objects has been significantly improved.
  \item We have invented a new network called HPC-Net. Many experiments have proved our method's progressiveness and superiority, ranking \textbf{first} in the KITTI Car 2D Object Detection Ranking. Achieve \textbf{97.59\%} average precision in moderate mode. On the KITTI Car 3D Object Detection Ranking, the overall ranking is currently fourth, and in hard mode, it ranks \textbf{first} (\textbf{82.65\%}). We also verified the generalizability of our method on the Waymo dataset.
\end{itemize}

\begin{figure*}[t]
    \centering
    \includegraphics[width=0.8\linewidth]{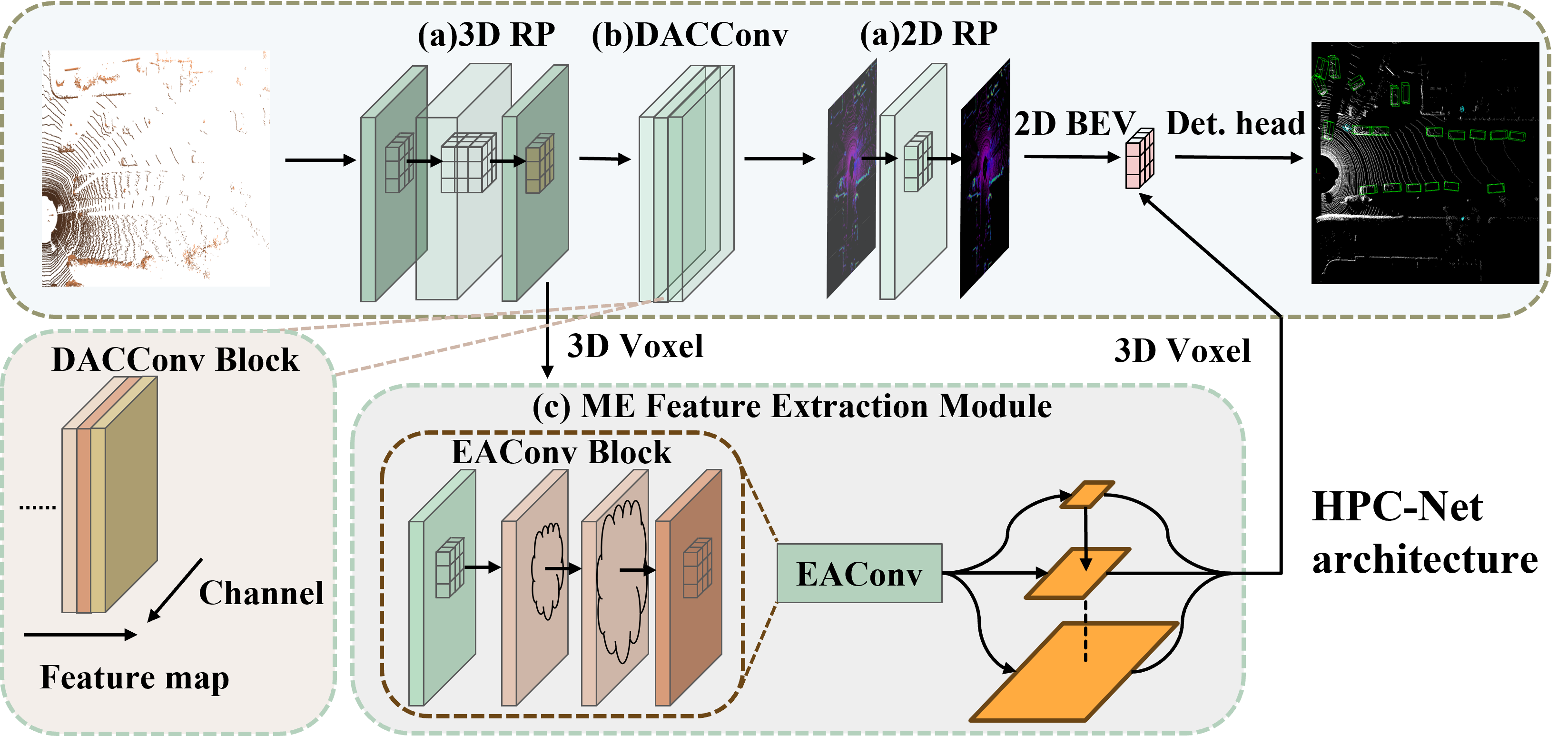}
    \caption{Overall architecture of HPC-Net.\ (a) Replaceable Pooling. 3D Replaceable Pooling elevates the voxel feature tensor by one dimension before pooling and then converts the tensor dimension into the initial input dimension. 2D Replaceable Pooling's principle is the same, but the difference is to convert 2D images into 3D models and then convert them back.\ (b) Depth Accelerated Convergence Convolution. The DACConv kernel is used to convolve voxel feature tensors from different channels and feature maps.\ (c) Multi-Scale Extended Receptive Field Feature Extraction Module. Extract 3D features of objects through multi-layer Deformable Convolution~\cite{Alpher27} and RROI (replaceable region of interest) pooling, and then output the results through multi-scale feature fusion network.}\label{figure1}
  \end{figure*}

%% file: 2_formatting.tex
\section{Related work}\label{sec:formatting}

\textbf{Object detection based on deep learning.} The field of object detection has been developing for over 20 years. As a crucial aspect of computer vision, it significantly impacts various domains, such as pedestrian detection, facial recognition, video analysis, and medical diagnosis. With the advancements in deep learning, object detection algorithms have greatly improved. Currently, deep-learning-based object detection algorithms can be categorized into one-stage~\cite{Alpher10,Alpher11} and two-stage object detection algorithms~\cite{Alpher12,Alpher13}. Moreover, there have been recent developments in object detection algorithms based on deep learning, including transformer-based methods~\cite{Alpher14,Alpher15} and voxel-based methods~\cite{Alpher03,Alpher31}. 
Our HPC-Net effectively enhances object detection accuracy, generalization, and robustness by increasing model depth and incorporating reasonable pooling steps as transitions between layers and modules.

\textbf{Object detection based on R-CNN.} R-CNN follows the principle of first generating region proposals from the original image, then resizing these proposals to a fixed size, classifying and scoring them, and finally applying NMS~\cite{Alpher16} and other techniques to filter out high-scoring bounding boxes for each category. Earlier methods, such as Fast R-CNN~\cite{Alpher12}, significantly reduced computational complexity by utilizing the spatial pyramid pooling (SPP) principle~\cite{Alpher35}. Faster R-CNN~\cite{Alpher13} further improved efficiency by introducing the region proposal network (RPN). PV-RCNN~\cite{Alpher04} combines point clouds with voxels to enhance accuracy. Recently, R-CNN-based methods have gained increasing attention in object detection. Our HPC-Net is currently a leading approach in R-CNN-based object detection.

\textbf{Voxel-based object detection.} Voxel-based methods, starting with Apple's VoxelNet~\cite{Alpher03}, organize unordered point cloud data into ordered high-dimensional feature data by converting point clouds into 3D voxel representations with consistent specifications. SECOND~\cite{Alpher30} and PV-RCNN~\cite{Alpher04} build upon this idea. Voxel R-CNN~\cite{Alpher31} proposes a two-stage detection framework consisting of a 3D backbone network, a 2D bird's-eye view proposal network, and detection heads, which has become a new mainstream approach, including methods like SFD~\cite{Alpher05}, TED~\cite{Alpher32}, and VirConv~\cite{Alpher23}. Our method is also based on the Voxel R-CNN~\cite{Alpher31} framework and surpasses previous models.

\textbf{Incorporating pooling methods for object detection.} Earlier approaches, such as PV-RCNN~\cite{Alpher04}, introduced ROI grid pooling, followed by Voxel R-CNN~\cite{Alpher31} incorporating Voxel RoI Pooling, and more recently, TED~\cite{Alpher32} introduced Transformation-Equivariant BEV Pooling based on maximum pooling and bilinear interpolation. However, these pooling methods are still based on maximum or average pooling and lack flexibility, leading to issues such as limited generalizability and accuracy. Therefore, we propose a novel pooling method called replaceable pooling.

\textbf{Multimodal 3D object detection.} Most state-of-the-art high-precision 3D object detection methods are multimodal, combining information from multiple sensors. Earlier methods, such as MV3D~\cite{Alpher28}, integrate LiDAR BV (Bird's Eye View), LiDAR FV (Front View), and RGB Image through a Region-based Fusion Network. 3D-CVF~\cite{Alpher17} divides multimodal data fusion into two stages, generating stronger Camera-LiDAR joint features in the first stage and aggregating encoded features and Camera-LiDAR joint features in the second stage. SFD~\cite{Alpher05} utilizes pseudo point clouds generated by depth completion to address fusion sparsity. 3D Grid-wise Attention Fusion (3D-GAF) is employed to mesh and fuse 3D RoI features in point cloud pairs. Recent methods like TED~\cite{Alpher32} and VirConv~\cite{Alpher23} utilize PENet~\cite{Alpher08} to integrate multimodal information. HPC-Net integrates 3D voxels and 2D BEV images, leveraging a novel feature extraction module called MEFEM.\@ This module increases the receptive field and integrates multi-scale features. Our approach achieves significantly improved accuracy across all modes of the KITTI dataset and effectively addresses the challenge of detecting highly occluded objects with high accuracy.

%% file: 3_finalcopy.tex
\section{HPC-Net for Object Detection}

This paper presents a novel object detection network, HPC-Net, which effectively enhances the detection accuracy and training convergence speed of existing object detection models. As illustrated in \cref{figure1}, our approach consists of three main components: (a) Replaceable Pooling, (b) Depth Accelerated Convergence Convolution, and (c) Multi-Scale Extended Receptive Field Feature Extraction Module.

We denote the element of the 3D tensor $\mathit{T^{n*h*w} \in \mathbb{R}^{N*H*W}}$ as $\mathit{T^{n*h*w}}$ ($\mathit{\mathbb{R}}$ represents all input tensors). Similarly, 4D tensors are represented. It is worth noting that, based on PyTorch's functionality, tensors can be reshaped arbitrarily while preserving the elements and values.

\subsection{Replaceable Pooling}

\begin{figure}[t]
    \centering
    \includegraphics[width=0.8\linewidth]{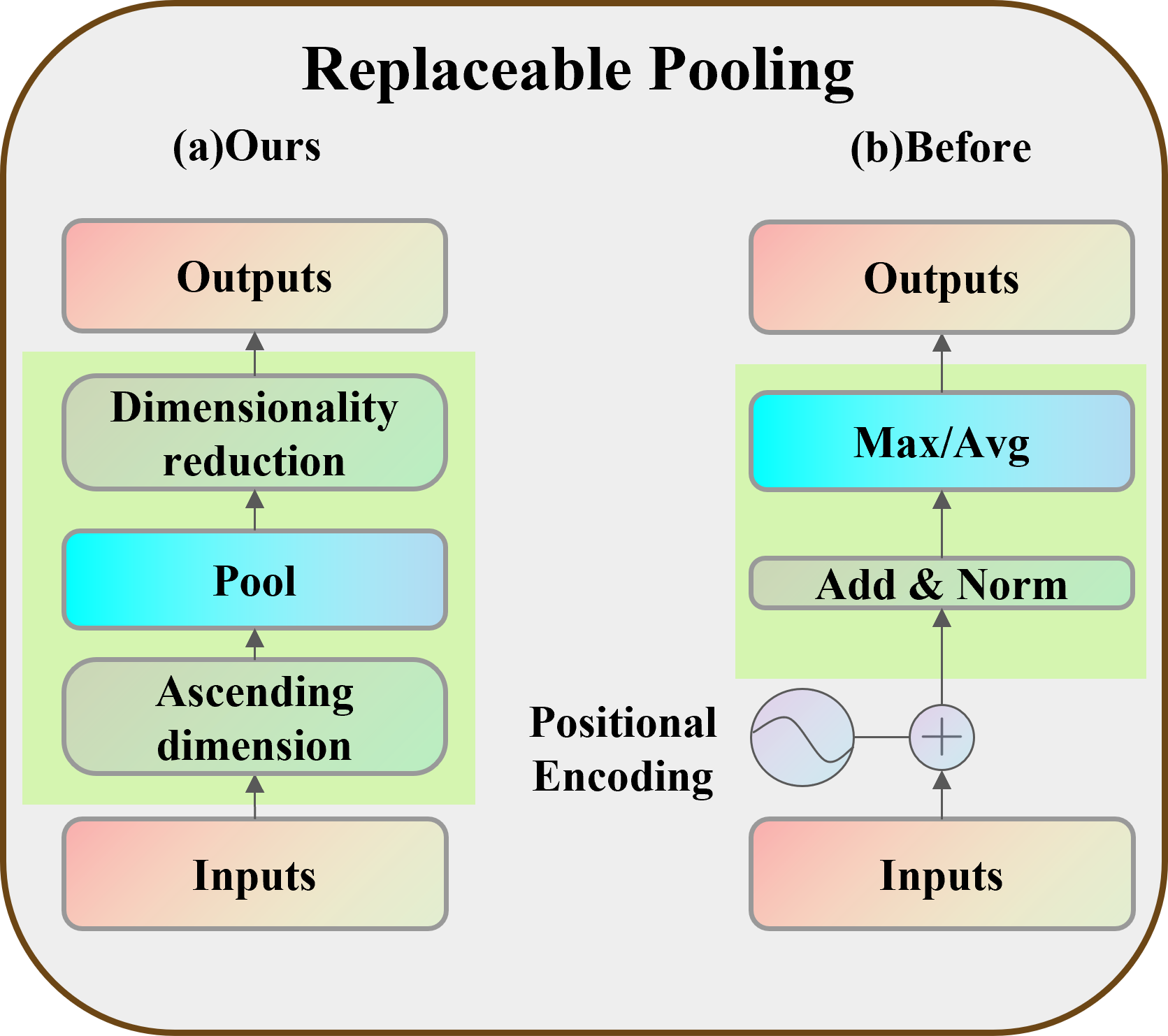}
    \caption{Replaceable Pooling.\ (a) Through dimensionality increase, pooling, and dimensionality reduction, these three steps pool input feature tensors in both 3D and 2D dimensions to improve accuracy and speed.\ (b) The previous method. Through comparison, it can be seen that our method is more concise and efficient in structure.}\label{figure2}
\end{figure}
  
Previously, the conversion of 3D point clouds to 2D BEV views did not involve a pooling step. However, some recent models have incorporated a combination of maximum pooling and bilinear interpolation~\cite{Alpher32}. In the Voxel Feature Encoding stage, traditional methods only computed the average or maximum value of points within voxels. In contrast, recent methods like TED~\cite{Alpher32} leverage the maximum pooling principle, where the maximum value of the internal point cloud set for each voxel is considered as the feature.

To illustrate the previous method and our approach, we will use 3D maximum pooling as an example.

The previous method first used slicing operations to calculate the sum of the feature $\mathit{f_{ij}}$ of all point clouds within a voxel. Secondly, perform normalization operations. Finally, by comparing the mapped values of this voxel after N rotations, the maximum value is used as the final output.
As shown in \cref{eq:1}, where $\mathit{F_k}$ represents the feature of the $\mathit{k_{th}}$ input voxel, $\mathit{f_{kij}}$ represents the feature of a single point cloud, $\mathit{n}$ represents the number of point clouds within a single voxel, and $\mathit{N}$ represents the number of voxel rotations (0 represents no rotation).

\begin{equation}  
 F_{k}=\max_{0\leq j\leq N}{|\frac{\sum_{i=1}^{n} f_{kij}}{n} |},k=1,2,\dots,\mathbb{R} .
\label{eq:1} 
\end{equation}

Replaceable Pooling consists of three steps applied to the voxel-based feature tensor: (1) Dimensionality increase $\mathit{\Psi (\cdot)}$; (2) Pooling $\mathit{\Phi (\cdot)}$; (3) Dimensionality reduction $\mathit{\varphi(\cdot)}$. 
As illustrated in \cref{eq:2}, 3D Replaceable Pooling transforms the tensor $\mathit{T^{n*c*h*w}_k \in \mathbb{R}^{N*C*H*W}}$ into $\mathit{\hat{T}^{n*c*d*h*w}_k \in \mathbb{R}^{N*C*D*H*W}}$, followed by pooling (\cref{eq:3}), and then converts the tensor $\mathit{T^{n*c*d*h*w} \in \mathbb{R}^{N*C*D*H*W}}$ back to $\mathit{T^{n*c*h*w} \in \mathbb{R}^{N*C*H*W}}$ (\cref{eq:4}). Our advantage lies in the independent nature of the pooling process, which allows users to modify pooling methods based on the requirements of different tasks. In our experiment, we employed eDSCW pooling~\cite{Alpher33}.

\begin{equation}  
    {
        \hat{T}^{n*c*d*h*w}_k =\Psi (T^{n*c*h*w}_k),k=1,2,\dots,\mathbb{R}.  
    }
\label{eq:2} 
\end{equation}

\begin{equation}  
   {
    F_{k i j}= \mathop{\Phi}\limits_{(p, q) \in \mathcal{R}_{i j}}(f_{k p q}),k=1,2,\dots,\mathbb{R}.
   }
\label{eq:3} 
\end{equation}

\begin{equation}  
    {
        T^{n*c*h*w}_k = \varphi(\hat{T}^{n*c*d*h*w}_k),k=1,2,\dots,\mathbb{R}.
    }
\label{eq:4} 
\end{equation}

As depicted in \cref{figure2}, our method RP exhibits a significantly simplified structure compared to previous methods, resulting in a notable advantage in terms of time complexity. The time complexity of our method is $\mathit{O (N \cdot (1+(h \cdot w)+1))}$, where $\mathit{h \cdot w}$ represents the entire feature range of the input feature tensor, $\mathit{N}$ represents the number of rotations. Every time the dimensionality is increased or decreased, it is $\mathit{O (1)}$ in terms of time complexity, as it only adds or subtracts a 1 to the shape element group of the tensor. In practical operation, $\mathit{h \cdot w}$ is equal to $\mathit{n}$ (number of point clouds within a voxel), $\mathit{N}$ is much smaller than $\mathit{n}$. Therefore, the time complexity of the Replaceable Pooling is $\mathit{O (N \cdot n)}$.
In contrast, the previous method had a time complexity of $\mathit{O (N \cdot (n+n+N))}$, which simplifies to $\mathit{O (2N \cdot n)}$. It is evident that our method exhibits half the time complexity of the previous method.

Analysis of spatial complexity. The previous method involved rotating $\mathit{O (N \cdot f (n))}$ ($\mathit{f (n)}$ is the pooling function), initializing $\mathit{O(n)}$, normalizing $\mathit{O(n)}$, and approximate pooling $\mathit{O(n)}$ to the maximum or average, resulting in a spatial complexity of $\mathit{O (3N \cdot n)}$. On the other hand, our method entails rotating $\mathit{O (N \cdot f (n))}$, initializing $\mathit{O (n)}$, and pooling $\mathit{O (n)}$, leading to a spatial complexity of $\mathit{O (2N \cdot n)}$. The memory usage remains unchanged as there are no new or reduced elements during tensor dimensionality increase or decrease. Theoretically, RP saves one-third of the spatial complexity compared to the original method.

\subsection{Depth Accelerated Convergence Convolution}

Traditional methods utilize standard convolution for feature extraction. However, a drawback of this approach is the slow convergence during training, which is not ideal for scenarios that require fast training. Depthwise Separable Convolution~\cite{Alpher07} has demonstrated that combining Depthwise Convolution and Pointwise Convolution can significantly reduce the parameter count of convolution operations, thereby accelerating the model's training process. DO-Conv~\cite{Alpher34} has proven that an over-parameterized structure similar to depth separable convolution can greatly improve convergence speed.

The usual method to solve the problem of slow convergence in object detection models is to increase the convolution speed by reducing the number of parameters, such as deep convolution. However, our experiments have shown that reducing the number of parameters leads to decreased accuracy despite increasing convergence speed during model training. To achieve faster convergence while maintaining accuracy, we have adopted two strategies for convolution operations. By combining channel-based convolution with feature-map-based convolution, we are able to achieve faster convergence without increasing computational complexity.

To illustrate the formula of DACConv, we choose to infer the position of a single element in the output feature map $\mathit{(x, y)}$.

\begin{equation}  
    {
        \begin{cases}
            O_{(x, y, C_{out})}= \sum\limits_{(x,y) \in K} [I (x ', y', C_{in}) \cdot K (\Delta x, \Delta y, C_{in})]\\
           \Delta x=x - x ', \Delta y=y - y'
        \end{cases}
    } 
\label{eq:5} 
\end{equation}

\cref{eq:5} represents multiplying all input feature maps by the corresponding convolutional kernels and adding them together to obtain the output feature map. Among them, $\mathit{C_{in}}$ represents the index of the input channel, $\mathit{C_{out}}$ represents the index of the output channel, $\mathit{x'}$ and $\mathit{y'}$ represent the coordinate offset on the input feature map, $\mathit{x}$ and $\mathit{y}$ represent the coordinates on the output feature map, $\mathit{I}$ represents the input feature map and $\mathit{I \in \mathbb{R}^{H_{in} * W_{in} * C_{in}}}$, $\mathit{K}$ represents the convolutional kernel and $\mathit{K \in \mathbb{R}^{K_{h} * K_{w} * C_{in}}}$, $\mathit{O}$ represents the output feature map and $\mathit{O \in \mathbb{R}^{H_{out} * W_{out} * C_{out}}}$.

The definition of convolutional kernel $\mathit{K}$ is determined by \cref{eq:6}. Among them, $\mathit{K_c \in \mathbb{R}^{D * (H * W) * C_{in}}}$ represents the channel based convolutional kernel, $\mathit{K_f \in \mathbb{R}^{C_{out} * (H * W) * C_{in}}}$ represents the input feature maps based convolutional kernel.

\begin{equation}  
    {
        K= K_c \cdot K_f 
    }
\label{eq:6} 
\end{equation}

\begin{figure}[t]
    \centering
    \includegraphics[width=0.8\linewidth]{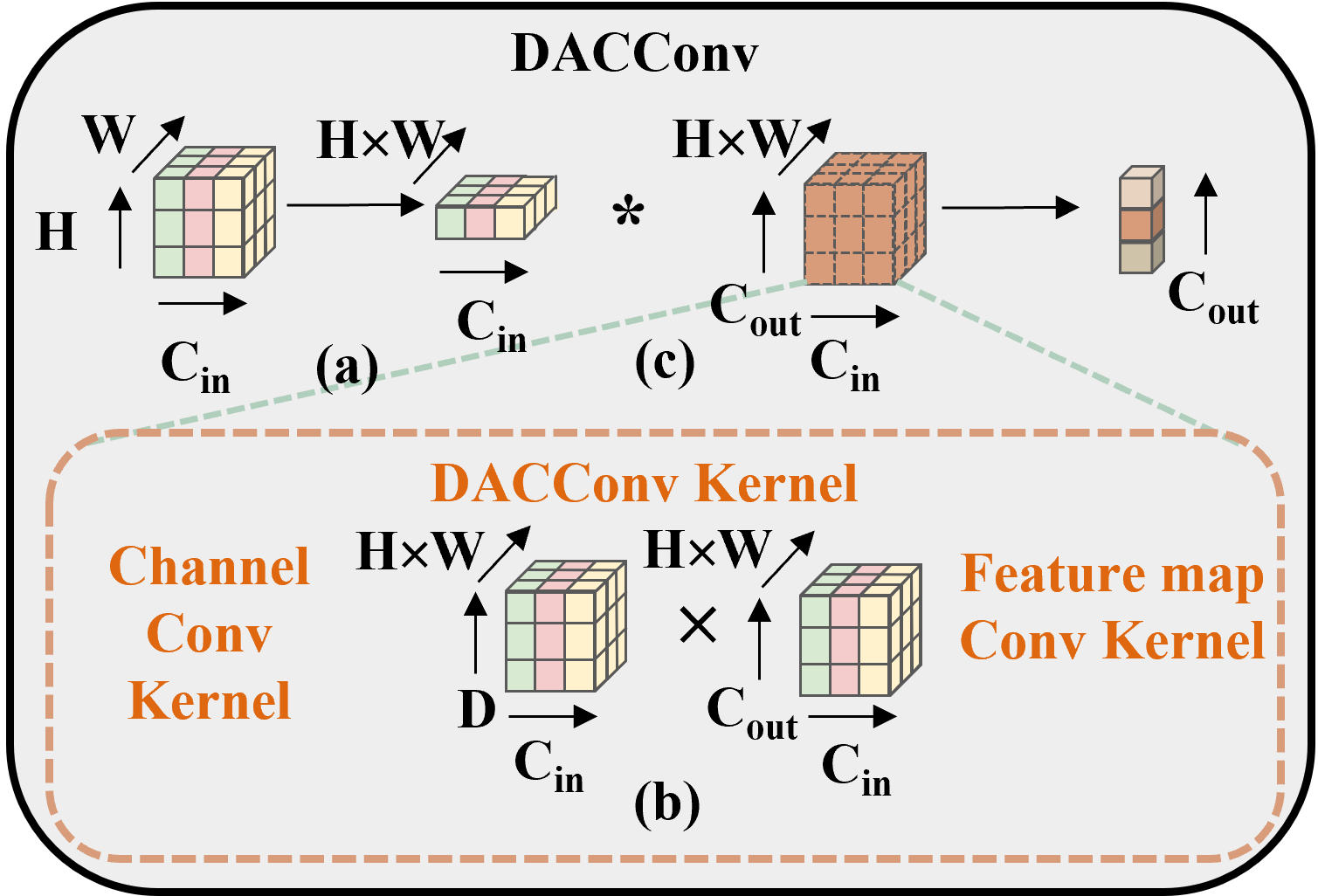}
    \caption{Depth Accelerated Convergence Convolution\. (a) Input tensor format as $\mathit{C_{in}*H*W}$. We are reshaping the input tensor as $\mathit{C_{in}*(H \times W)}$.\ (b) Multiply the channel convolution kernel $\mathit{(C_ {in} * D  * H  *W)}$ by the feature map convolution kernel $\mathit{(C_ {in} *  C_ {out} * H  * W)}$ to generate a DACConv kernel\. (c) Convolve input tensor $\mathit{(C_{in} * (H \times  W))}$ with DACConv Kernel $\mathit{(C_{in} * C_{out} * (H \times W))}$. $\mathit{C_{in}}$ indicates the index of the input channel. $\mathit{C_{out}}$ indicates the index of the output channel.}\label{figure3}
\end{figure}

The specific operation is illustrated in \cref{figure3}. Initially, we assign a distinct convolution kernel to each input channel. Subsequently, we employ a shared convolution kernel for different input feature maps. Finally, we multiply these two convolution kernels to obtain a new convolution kernel, which we refer to as the DACConv convolution kernel. By combining these two strategies, DACConv can achieve more learning dimensions while maintaining the same inference time complexity, thereby effectively improving the training convergence speed of the model. Our experiments also validate this improvement.

Regarding the time complexity, let's consider a single convolutional layer as an example. The time complexity of previous standard convolutional neural networks was $\mathit{O(M \cdot m \cdot P \cdot Q \cdot C_{in} \cdot C_{out})}$. Here, $\mathit{M}$ represents the length of the output feature map, $\mathit{m}$ represents the width of the output feature map, $\mathit{P}$ represents the length of the input feature map, $\mathit{Q}$ represents the width of the input feature map, $\mathit{C_{in}}$ represents the number of input channels, and $\mathit{C_{out}}$ represents the number of output channels.
Since DACConv is consistent with standard convolution in terms of input and output, the time complexity of our method for reasoning is also $\mathit{O (M \cdot m \cdot P \cdot Q \cdot C_{in} \cdot C_{out})}$.
During practical operation, it can be observed that DACConv introduces an additional step of synthesizing a convolutional kernel into the standard convolution process. Consequently, the actual running time only increases by one tensor multiplication of $\mathit{O (D \cdot C_{out})}$. The additional computational complexity of DACConv within the entire convolutional neural network framework is $\mathit{O (\sum_{l=1}^{D} D \cdot C_{l})}$. Compared to the time complexity of the entire convolutional neural network $\mathit{O\left(\sum_{l=1}^{D} M_{l}\cdot m_{l} \cdot P_{l} \cdot Q_{l} \cdot C_{l-1} \cdot C_{l}\right)}$ ($\mathit{D}$ represents depth, $\mathit{l}$ represents the current convolutional layer), the increased computational time is negligible.

\subsection{Multi-Scale Extended Receptive Field Feature Extraction Module}
Previous voxel-based methods utilized the Pointnet module~\cite{Alpher24} for 3D feature extraction. Pointnet++~\cite{Alpher25} introduced a multi-level architecture that effectively extracts both local and global features. Voxel-RCNN~\cite{Alpher31} proposed the use of Voxel ROI Pooling to extract adjacent voxel features in proposals.

\begin{figure}[t]
    \centering
    \includegraphics[width=0.8\linewidth]{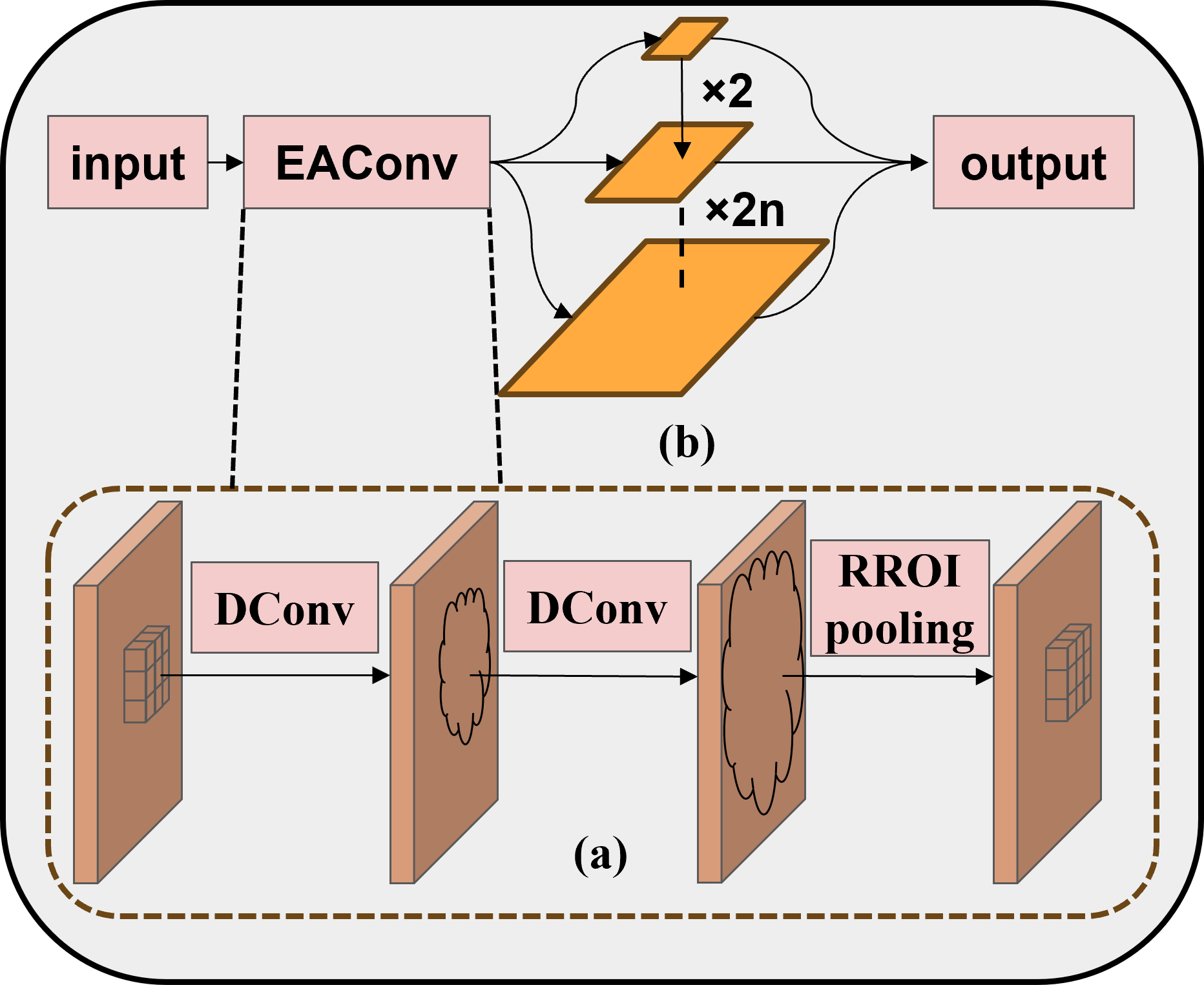}
    \caption{Multi-Scale Extended Receptive Field Feature Extraction Module.\ (a) Extending Area Convolution. The EAConv block consists of two layers of Dconv and RROI pooling, greatly increasing the receptive field area. Dconv representative Deformable Convolution~\cite{Alpher27}.\ (b) Multi-scale Feature Fusion Network. The input is the output of EAConv, fusing features from multiple scales as the output.}\label{figure4}
\end{figure}

To address the challenging task of detecting occluded objects in object detection, we have developed a novel feature extraction module called the Multi-Scale Extended Receptive Field Feature Extraction Module. As depicted in \cref{figure4}, this module consists of two parts. The first part, named EAConv, expands the receptive field through multi-layer Deformable Convolution~\cite{Alpher27} and further extracts object features using RROI pooling. The second part employs a multi-scale fusion strategy to combine the output results of the initial pooling step across multiple scales, thereby enhancing the module's feature expression capability.

In EAConv, our approach to improving occluded object detection involves increasing the depth of convolutional neural networks and utilizing deformable convolution~\cite{Alpher27} to expand the receptive field of the convolutional layer. We enhance the standard convolution by employing a multi-layer deformable convolution~\cite{Alpher27} with an RROI pooling architecture. The formula for EAConv is defined as follows (\cref{eq:7}), where $\mathit{p_{0}}$ is each point in the output feature map, corresponding to the center point of the convolution kernel; $\mathit{p_{n}}$ is each offset of $\mathit{p_{0}}$ within the convolutional kernel range; $\mathit{w(p_{n})}$ is the weight of the corresponding position of the convolutional kernel; $\mathit{x(p_{0}+p_{n})}$ is the element value at the position of $\mathit{p_{0}+p_{n}}$ on the input feature map, with each point introducing an offset $\mathit{\Delta p_{n}}$. $\mathit{\Phi (\cdot)}$ represents RROI pooling. $\mathit{O(p_{0})}$ indicates that the output of EAConv undergoes two layers of deformable convolution~\cite{Alpher27} followed by RROI pooling.

\begin{equation}  
    {
        \begin{cases}
         y\left(p_{0}\right)= \sum\limits_{p_{n} \in R}w\left(p_{n}\right) \cdot x\left(p_{0}+p_{n}+\Delta p_{n}\right)\\
         O\left(p_{0}\right)=\Phi \left(y\left( y\left(p_{0}\right)\right)\right) 
        \end{cases}
    }
\label{eq:7} 
\end{equation}

Deformable Convolution~\cite{Alpher27} achieves a larger receptive field by incorporating offsets into the convolution kernel and combining it with the input. Our experiments have demonstrated that the highest accuracy is achieved when the deformable convolution~\cite{Alpher27} is increased to two layers.

Leveraging the concept of Replaceable Pooling (RP), we introduce a new interface for the existing ROI pooling module in current object detection methods. This new module, called RROI pooling, is not limited to only two pooling methods, namely maximum pooling and average pooling. Users of the new ROI pooling module can now choose the pooling method that best suits their requirements.

After EAConv convolutes the input feature tensor, we progressively expand the pooling area of RROI pooling in a 2-fold proportional sequence. This generates a series of feature output maps with dimensions equal to twice the proportional sequence. The principle of multi-scale fusion is illustrated in \cref{eq:8}, where $\mathit{k}$ represents the layer number of the feature map, $\mathit{w}$ and $\mathit{h}$ respectively represent the width and height of the object box, and $\mathit{k_0}$ is a hyperparameter. By fusing these feature maps, we achieve multi-scale feature fusion results, effectively enhancing the performance of the feature extraction module.

\begin{equation}  
    {
        k=\lfloor k_0+\log_{2}(\sqrt{w \cdot h} / 224)\rfloor
    }
\label{eq:8} 
\end{equation}

\begin{table*}
    \centering
    \begin{tabular}{@{}l|c|c|c@{}}
    \bottomrule
    Method &
    Modality &
    \begin{tabular}{ccc} &Car 3D (R40)& \\Easy&Mod.&Hard\end{tabular} & 
    \begin{tabular}{ccc} &Car 2D (R40)& \\Easy&Mod.&Hard\end{tabular} \\
    \hline
    PV-RCNN (2020)~\cite{Alpher04} &
    LiDAR &
    90.25\ \ \qquad  81.43 \ \ \qquad  76.82 &
    98.17\ \ \qquad  94.70 \ \ \qquad  92.04 \\
    Voxel-RCNN (2021)~\cite{Alpher31} &
    LiDAR &
    90.90\ \ \qquad  81.62 \ \ \qquad  77.06 &
    96.49\ \ \qquad  95.11 \ \ \qquad  92.45 \\
    CT3D (2021)~\cite{Alpher22} &
    LiDAR &
    87.83\ \ \qquad  81.77 \ \ \qquad  77.16 &
    96.28\ \ \qquad  93.30 \ \ \qquad  90.58 \\
    SE-SSD (2021)~\cite{Alpher20} &
    LiDAR &
    91.49\ \ \qquad  82.54 \ \ \qquad  77.15 &
    96.69\ \ \qquad  95.60 \ \ \qquad  90.53 \\
    TED (2023)~\cite{Alpher32} &
    LiDAR+RGB &
    91.61\ \ \qquad  85.28 \ \ \qquad  80.68 &
    96.64\ \ \qquad  96.03 \ \ \qquad  93.35 \\
    LoGoNet (2023)~\cite{Alpher21} &
    LiDAR+RGB &
    91.80\ \ \qquad  85.06 \ \ \qquad  80.74 &
    96.60\ \ \qquad  95.55 \ \ \qquad  93.07 \\
    SFD (2022)~\cite{Alpher05} &
    LiDAR+RGB &
    91.73\ \ \qquad  84.76 \ \ \qquad  77.92 &
    \textbf{98.97}\ \ \qquad  96.17 \ \ \qquad  91.13 \\
    3D HANet (2023)~\cite{Alpher41} &
    LiDAR+RGB &
    90.79\ \ \qquad  84.18 \ \ \qquad  77.57 &
    98.61\ \ \qquad  95.73 \ \ \qquad  92.96 \\
    CasA++ (2022)~\cite{Alpher40} &
    LiDAR+RGB &
    90.68\ \ \qquad  84.04 \ \ \qquad  79.69 &
    95.83\ \ \qquad  95.28 \ \ \qquad  \textbf{94.28} \\
    GLENet-VR (2022)~\cite{Alpher43} &
    LiDAR &
    91.67\ \ \qquad  83.23 \ \ \qquad  78.43 &
    96.85\ \ \qquad  95.81 \ \ \qquad  90.91 \\
    VPFNet (2022)~\cite{Alpher42} &
    LiDAR+RGB &
    91.02\ \ \qquad  83.21 \ \ \qquad  78.20 &
    96.64\ \ \qquad  96.15 \ \ \qquad  91.14 \\
    \hline
    HPC-Net (ours) &
    LiDAR+RGB &
    \textbf{92.08}\ \ \qquad  \textbf{85.50} \ \ \qquad  \textbf{82.65} &
    98.61\ \ \qquad  \textbf{97.59} \ \ \qquad  93.01 \\
    \toprule
    \end{tabular}
    \caption{3D and 2D detection results on the KITTI test set. The best methods are in bold. HPC-Net performs best in 3D object detection results for all modes and 2D object detection results for moderate mode.}\label{table1}
\end{table*}

\section{Experiment}

\subsection{Dataset and evaluation criteria}

\textbf{KITTI Dataset and evaluation criteria.} To demonstrate the superiority and progressiveness of HPC-Net, we integrated it into two state-of-the-art 3D object detection frameworks, VirConv~\cite{Alpher23} and TED~\cite{Alpher32}, both of which have been verified using the KITTI dataset~\cite{Alpher09}. The KITTI 3D object detection dataset is divided into training and testing sets. The training set consists of 7481 LiDAR and image frames, while the testing set consists of 7518 LiDAR and image frames. Following the requirements of VirConv~\cite{Alpher23} and TED~\cite{Alpher32}, we further divided the training data into a training segmentation of 3712 frames and a validation segmentation of 3769 frames.

We adopted the widely recognized evaluation metric, 3D average accuracy (AP), under 40 recall thresholds (R40). For cars, the IoU threshold for this metric is set to 0.7. Additionally, we utilized the KITTI odometer dataset, which contains 43552 LiDAR and image frames, as a large-scale unlabeled dataset for the odometer task. Specifically, we uniformly sampled 10888 frames (represented as a semi-dataset) and used them to train our VirConv-S-based HPC-Net.

\textbf{Waymo Dataset and evaluation criteria.} The WOD~\cite{Alpher29} dataset consists of 798 training sequences and 202 validation sequences. The official evaluation metrics include average accuracy (mAP) for L1 and L2 difficulty levels, as well as titles (mAPH and APH) for the same difficulty levels.

\subsection{Setup Details}

We named our HPC-Net variants based on the corresponding backbone models: HPC-VT (based on VirConv-T~\cite{Alpher23}), HPC-VS (based on VirConv-S~\cite{Alpher23}), and HPC-T (based on TED-M~\cite{Alpher32}). Our experimental settings are consistent with the baseline to ensure fairness.

\textbf{Network details.} All HPC-Net models adopt an architecture based on the Voxel-RCNN~\cite{Alpher31} backbone and utilize PENet~\cite{Alpher08} to generate virtual points.

\textbf{Loss and data augmentation.}All HPC-Net models use the same training loss as TED~\cite{Alpher32} and VirConv~\cite{Alpher23}. We employ commonly used techniques for data augmentation, including ground live sampling, local transformations (rotation and translation), and global transformations (rotation and flipping).

\textbf{Training details.} We train the models using 8 Tesla V100 GPUs in parallel. We also support training with a single or multiple 3090 GPUs. The learning rate is set to 0.01, and a single-cycle learning strategy is employed. HPC-VT is trained for 60 cycles, HPC-VS for five cycles, and HPC-T for 30 cycles.

\subsection{Main Results}

\textbf{KITTI test set.} To demonstrate the superiority of our method and maintain an objective and impartial approach, we submitted our HPC-Net to the official KITTI Test dataset website. The results, shown in \Cref{table1}, indicate that our method ranks first in Car 2D (R40) detection results and significantly outperforms other methods in Car 3D (R40) detection results over the past four years. Overall, our method ranks fourth on the official KITTI Car 3D (R40) dataset and first in hard mode.

\textbf{KITTI validation set.} To showcase the universal applicability and superiority of our method, we validated it on the KITTI validation set using TED~\cite{Alpher32} and VirConv~\cite{Alpher23} as baselines, which are the top two publicly available methods in the KITTI dataset's object detection ranking list. As shown in \Cref{table2}, our method outperforms the baselines in Car 3D (R40), and it also surpasses the baselines in Car 2D (R40) for both moderate and hard modes.

\textbf{Waymo validation set.} For additional results and details, please refer to the supplementary materials.

\begin{table*}[t]
    \centering
    \begin{tabular}{@{}l|c|c|c@{}}
    \bottomrule
    Method &
    Modality &
    \begin{tabular}{ccc} &Car 3D (R40)& \\Easy&Mod.&Hard\end{tabular} & 
    \begin{tabular}{ccc} &Car 2D (R40)& \\Easy&Mod.&Hard\end{tabular} \\
    \hline
    TED-S &
    LiDAR &
    92.68\ \ \qquad  87.61 \ \ \qquad  85.62 &
    98.55\ \ \qquad  97.02 \ \ \qquad  94.97 \\
    TED-M &
    LiDAR+RGB &
    95.46\ \ \qquad  88.93 \ \ \qquad  84.45 &
    99.41\ \ \qquad  96.16 \ \ \qquad  93.66 \\
    HPC-T &
    LiDAR+RGB &
    95.73\ \ \qquad  89.06 \ \ \qquad  86.48 &
    99.45\ \ \qquad  96.13 \ \ \qquad  93.54 \\
    \hline
    VirConv-T &
    LiDAR+RGB &
    94.36\ \ \qquad  88.17 \ \ \qquad  85.61 &
    \textbf{99.50}\ \ \qquad  97.70 \ \ \qquad  95.33 \\
    HPC-VT &
    LiDAR+RGB &
    95.47\ \ \qquad  89.51 \ \ \qquad  87.68 &
    99.31\ \ \qquad  97.47 \ \ \qquad  95.13 \\
    \hline
    VirConv-S &
    LiDAR+RGB &
    95.31\ \ \qquad  88.51 \ \ \qquad  88.65 &
    99.03\ \ \qquad  97.69 \ \ \qquad  95.48 \\
    HPC-VS &
    LiDAR+RGB &
    \textbf{95.96}\ \ \qquad  \textbf{90.98} \ \ \qquad  \textbf{89.00} &
    99.22\ \ \qquad  \textbf{97.77} \ \ \qquad  \textbf{95.52} \\
    \toprule
    \end{tabular}
    \caption{3D and 2D detection results on the KITTI valuation set, the best methods are in bold.}\label{table2}
\end{table*}

\begin{table*}
\centering
\begin{tabular}{@{}l|c|c|c|c|c@{}}
\bottomrule
Method &
3D Pooling Method &
2D Pooling Method &
\begin{tabular}{c} Car3D (R40) \\EasyMod.Hard\end{tabular} & 
\begin{tabular}{c} Car2D (R40) \\EasyMod.Hard\end{tabular} &
Time (ms) \\
\hline
TED-S &
- &
- &
92.68 87.61 \textbf{85.62} &
98.55 97.02 \textbf{94.97} &
166 \\
TED-S+RP &
eDSCW &
eM &
92.54 87.25 85.24 &
98.80 96.91 94.90 &
\textbf{79} \\
TED-S+RP &
eDSCW &
eDSCW &
\textbf{93.21} \textbf{87.76} 85.37 &
\textbf{99.04} \textbf{97.04} 94.83 &
251 \\
\toprule
\end{tabular}
\caption{The KITTI valuation set's 3D and 2D Replaceable Pooling detection results are displayed in bold for the best method.}\label{table3}
\end{table*}

\subsection{Ablation Experiment}

To ensure the fairness of the experiment, we independently validated each part of HPC-Net on the KITTI validation set.

\textbf{Applying Replaceable Pooling only.} We conducted experiments using TED-S~\cite{Alpher32} as the baseline and performed multiple comparative experiments to demonstrate the effectiveness of Replaceable Pooling. As shown in \Cref{table3} (where the second column represents 3D Pooling and the third column represents 2D Pooling), we concluded that the second combination yielded the fastest results. By slightly sacrificing the accuracy of 2D and 3D detection, we achieved a significant improvement in computational speed, resulting in an efficiency boost of approximately 52.4\%. The third combination achieved the highest accuracy, with both 2D and 3D detection accuracy being the highest and showing significant improvement.

\begin{figure}[t]
    \centering
    \includegraphics[width=0.9\columnwidth]{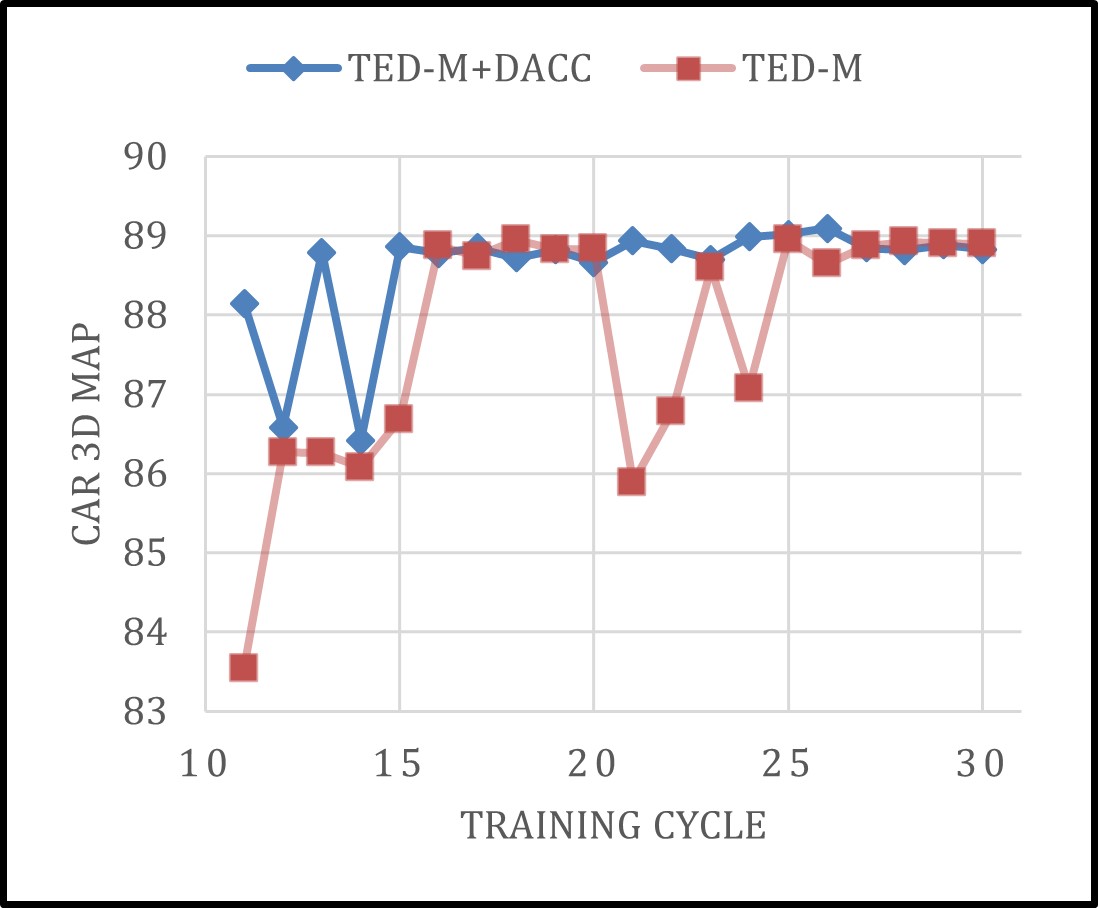} 
    \caption{Apply only Depth Accelerated Convergence Convolution. After applying Depth Accelerated Convergence Convolution, TED-M reached convergence at the 15th cycle. Before applying Depth Accelerated Convergence Convolution, TED-M reaches convergence at the 27th cycle.}\label{figure5}
\end{figure}

\textbf{Applying Depth Accelerated Convergence Convolution only.} We conducted experiments using TED-M~\cite{Alpher32} as the baseline to demonstrate the effectiveness of Depth Accelerated Convergence Convolution. We compared the mean average precision (mAP) of TED-M using Depth Accelerated Convergence Convolution with TED-M over 30 training cycles. Since neither method converged in the first ten training cycles and the accuracy fluctuated significantly, we excluded the results of the first ten training cycles. As depicted in \cref{figure5}, after applying DACConv, TED-M reached a convergence state in the fifteenth cycle, whereas without DACConv, TED-M only reached convergence in the 27th cycle. The convergence speed increased by approximately 44.4\%. Additionally, TED-M+DACConv achieved better accuracy than TED-M.

\textbf{Applying Multi-Scale Extended Receptive Field Feature Extraction Module only.} As shown in \Cref{table4}, we conducted experiments using TED-M as the baseline to demonstrate the effectiveness of MEFEM\@. By applying MEFEM, we compared the 2D and 3D object detection accuracy of baseline TED-M with TED-M+MEFEM\@. All the results mentioned above are from the KITTI validation set of the same device. Notably, all the data of MEFEM outperformed the baseline.

\begin{table}
    \centering
    \begin{tabular}{@{}lcc@{}}
        \toprule
        Method &
        \begin{tabular}{c} Car3D (R40) \\EasyMod.Hard\end{tabular} &
        \begin{tabular}{c} Car2D (R40) \\EasyMod.Hard\end{tabular} \\
        \midrule
        TED-M  &
        95.46 88.93 84.45 &
        99.41 96.16 93.66 \\ 
        \midrule
        TED+MEFEM  &
        \textbf{95.80} \textbf{89.22} \textbf{84.59} & 
        \textbf{99.48} \textbf{96.28} \textbf{93.70} \\ 
        \bottomrule
    \end{tabular}
    \caption{Apply only MEFEM\@. TED-M+MEFEM is much better than TED-M in any mode. The best methods are in bold.}\label{table4}
\end{table}

\section{Conclusion}

This paper introduces a novel object detection network, HPC-Net, designed for multimodal two-stage 3D object detection. By incorporating Replaceable Pooling, Depth Accelerated Convergence Convolution, and Multi-Scale Extended Receptive Field Feature Extraction Module, we have achieved significant improvements in the accuracy of detecting 2D and 3D objects, as well as the convergence speed of the models. Additionally, we have opened up the pooling layer interface to cater to different task scenarios. Our HPC-Net currently holds the top positions in the KITTI Car 2D and 3D Test Rankings, ranking first and fourth, respectively. Notably, in the hard mode, our 3D car detection test ranks first. These results serve as strong evidence of the effectiveness and progressive nature of our proposed method.

\clearpage